\documentclass[10pt,twocolumn,letterpaper]{article}

\usepackage{multirow}  % 引入 multirow 包用于跨行
\usepackage{geometry}  % 用于调整页面边距
\usepackage{array}     % 用于自定义列宽
\usepackage{graphicx}  % 用于表格字体大小控制
\usepackage{caption} 
\usepackage{amsmath}
\usepackage{amssymb}
\usepackage{mathtools}
\usepackage{amsthm}
\usepackage{pifont}    % 提供 \ding{} 以生成打勾和叉号
\usepackage{algorithm}
\usepackage{algpseudocode}
\usepackage{makecell}

\usepackage[accsupp]{axessibility}

\usepackage{xcolor}
\usepackage{pifont}

\newcommand{\cmark}{\textcolor{green!60!black}{\ding{51}}}
\newcommand{\xmark}{\textcolor{red!70!black}{\ding{55}}}

\usepackage{cvpr}              % To produce the CAMERA-READY version
\usepackage[normalem]{ulem}

\definecolor{cvprblue}{rgb}{0.21,0.49,0.74}
\usepackage[pagebackref,breaklinks,colorlinks,allcolors=cvprblue]{hyperref}

\def\paperID{*****} % *** Enter the Paper ID here
\def\confName{CVPR}
\def\confYear{2026}

\title{LongStream: Long-Sequence Streaming Autoregressive Visual Geometry}
\author{
Chong Cheng$^{1,2}$ \quad
Xianda Chen$^{4}$ \quad
Tao Xie$^{2,3}$ \quad
Wei Yin$^{2}$ \\
Weiqiang Ren$^{2}$ \quad
Qian Zhang$^{2}$ \quad
Xiaoyang Guo$^{2,\ddagger}$ \quad
Hao Wang$^{1,\dagger}$ \\
\vspace{2pt}
\small
$^{1}$HKUST (GZ)\quad
$^{2}$Horizon Robotics\quad
$^{3}$ZJU \quad
$^{4}$CSU \\
\footnotesize
$^{\ddagger}$Project Lead \quad
$^{\dagger}$Corresponding Author
}

\begin{document}
\twocolumn[{
\maketitle
\begin{center}
\vspace{-31pt}
    \includegraphics[width=0.88 \textwidth]{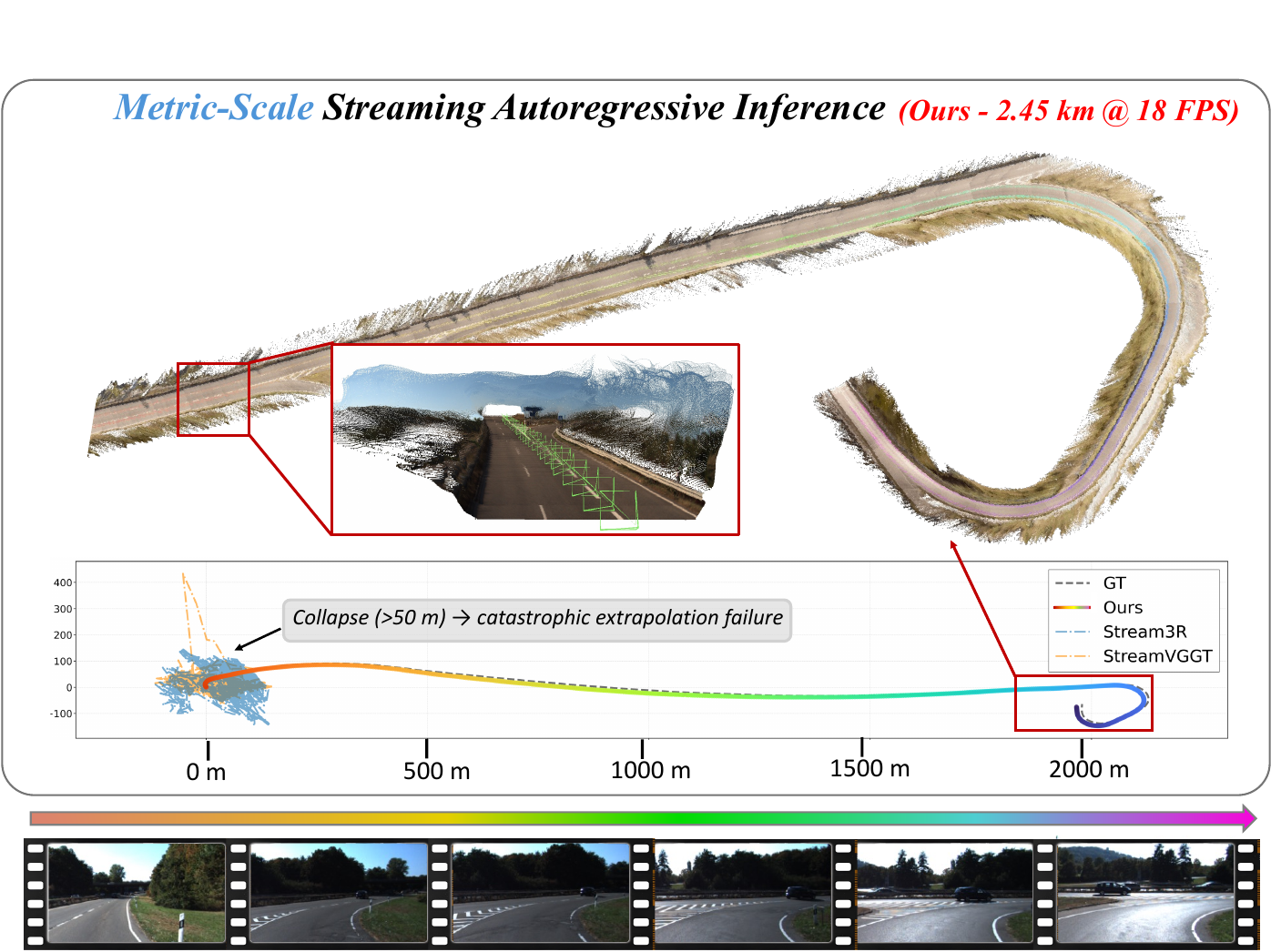}
    \label{fig:teaser}
    \vspace{-7pt}
    \captionof{figure}{ \textbf{
    Streaming Autoregressive Model Comparison for Metric-Scale Scene Reconstruction.} Existing streaming models (\eg, Stream3R~\cite{lan2025stream3r}, StreamVGGT~\cite{zhuo2025streaming}) collapse within tens of meters, suffering from extrapolation errors. In contrast, our proposed LongStream delivers stable, kilometer-scale reconstruction. Its gauge-decoupled formulation and cache-consistent training preserve metric accuracy and geometric stability, sustaining 18 FPS performance across multi-kilometer sequences.    
}
\end{center}
}]

\begin{abstract}
Long-sequence streaming 3D reconstruction remains a significant open challenge. 
Existing autoregressive models often fail when processing long sequences because they anchor poses to the first frame, leading to attention decay, scale drift, and extrapolation errors.
We introduce LongStream, a novel gauge-decoupled streaming visual geometry model for metric-scale scene reconstruction across thousands of frames under a strictly online, future-invisible setting.
Our approach is threefold. First, we discard the first-frame anchor and predict keyframe-relative poses. This reformulates long-range extrapolation into a constant-difficulty local task. Second, we introduce orthogonal scale learning. This method fully disentangles geometry from scale estimation to suppress drift. 
Finally, we identify attention bias issues in Transformers, including attention-sink reliance and long-term KV-cache saturation. 
We propose cache-consistent training combined with periodic cache refresh. This approach suppresses attention biases and contamination over ultra-long sequences and reduces the gap between training and inference.
Experiments show that LongStream achieves state-of-the-art performance, enabling stable, metric-scale reconstruction over kilometer-scale sequences at 18 FPS. Project Page: \url{https://3dagentworld.github.io/longstream/}

\end{abstract}    
\section{Introduction}
\label{sec:intro}
Geometry reconstruction, the joint estimation of camera poses and dense 3D structure from image sequences~\cite{Schonberger_2016_CVPR,schoenberger2016mvs,cheng2025graphguidedscenereconstructionimages,yao2018mvsnetdepthinferenceunstructured,song2024gvkfgaussianvoxelkernel, 10.1145/3664647.3681168}, is a cornerstone technology for applications like autonomous driving, AR/VR, and embodied robotics~\cite{10.1145/3664647.3681641,shen2025adhmraligningdiffusionbasedhuman}. These domains demand systems that can robustly process long-sequence video streams in real time. 

\begin{figure}
  \begin{center}
  \includegraphics[width=\linewidth]{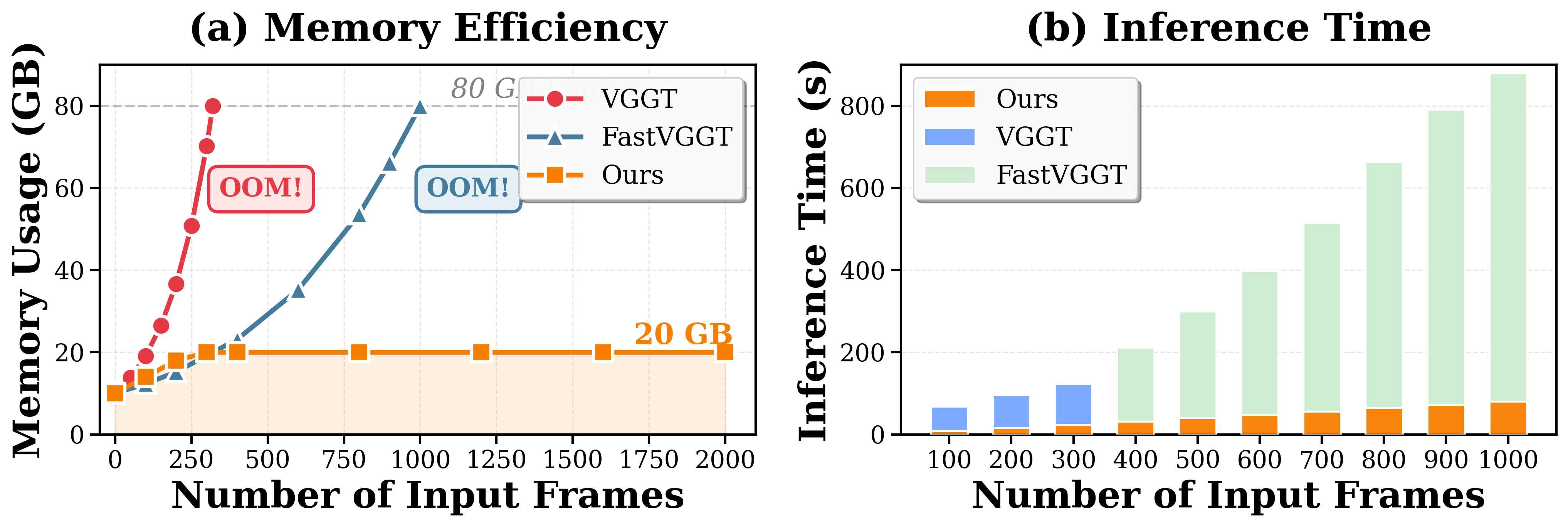}
  \end{center}
  \vspace{-18pt}
  \caption{\textbf{Memory and runtime comparison.} Our method keeps memory and latency stable, whereas VGGT and FastVGGT grow rapidly and hit OOM on long sequences.}
  \label{fig:memory}
  \vspace{-15pt}
\end{figure}

Conventional pipelines~\cite{Schonberger_2016_CVPR,schoenberger2016mvs,yao2018mvsnetdepthinferenceunstructured,yao2019recurrent,Wang_2024} and recent Transformer-based approaches~\cite{wang2024dust3r,leroy2024grounding,wang2025vggt,wang2025pi,hu2025vggt4dminingmotioncues,wang2025jasmine} achieve state-of-the-art accuracy, but are inherently offline. They require reprocessing the entire sequence to integrate a new frame~\cite{lan2025stream3r, zhuo2025streaming}, leading to massive computational redundancy and precluding real-time use.
To address this, streaming models have been proposed. Recent works such as STream3R~\cite{lan2025stream3r} and StreamVGGT~\cite{zhuo2025streaming} employ causal Transformers and KV-caching to build reconstructions incrementally. However, these models suffer from catastrophic extrapolation failure when processing long sequences. As shown in Figures 1 and ~\ref{fig:memory}, existing streaming methods incrementally reconstruct scenes in linear time, but their trajectories collapse within tens of meters, leading to complete tracking failure.

We argue that this failure stems from the \textbf{``gauge-coupled''} design inherent in current models. They are anchored to the first-frame coordinate system and trained to regress absolute poses. 
This forces the model to learn a position-fixed mapping, making long-sequence prediction increasingly difficult. Due to hardware constraints, the model only sees small indices in a training batch, making it difficult to extrapolate reliably to large indices at inference. As a result, a domain gap emerges under the “train-short, test-long” bias~\cite{press2022trainshorttestlong,dai2019transformerxlattentivelanguagemodels}.

To this end, we propose LongStream, a \textbf{``gauge-decoupled''} streaming autoregressive geometry framework. The framework theoretically decouples the gauge freedoms of global $SE(3)$ coordinates and metric scale.

First, to handle coordinate gauge over long sequences, we remove the fixed first-frame anchor. Instead, we regress keyframe-relative poses. This reformulates the ill-posed long-range extrapolation problem into a constant-difficulty local estimation task. As a result, the predictions become invariant to the global coordinate choice.

Second, to tackle Sim(3) scale drift, we introduce an orthogonal scale learning mechanism~\cite{yang2024depthanythingunleashingpower}. We decouple geometry learning from metric scale estimation at the objective level. The geometry branch optimizes shape in a scale-invariant space, while a dedicated scale head independently predicts the global scale factor. This effectively reduces scale entanglement and ensures stable metric outputs.

Beyond this gauge-decoupled formulation, the streaming Transformer architecture itself presents challenges. We identify that attention sink~\cite{xiao2024efficientstreaminglanguagemodels}, i.e., its biased dependence on the first-frame token, and long-term KV-cache contamination are primary causes of temporal degradation and drift.
To address this, we propose a cache-consistent training scheme. It aligns training and inference contexts by explicitly passing and trimming the cache during training. We further introduce a periodic cache refresh approach, which marginalizes stale context, mitigates long-term memory saturation, and stabilizes geometry.

Experiments on both outdoor (KITTI, vKITTI, Waymo) and indoor (TUM-RGBD, ETH3D, 7Scenes) datasets show that LongStream achieves state-of-the-art streaming reconstruction. It enables \textbf{real-time (18 FPS), metric-scale} AR reconstruction over kilometer-scale sequences. Our contributions are summarized as follows:

\begin{itemize}
    \item We propose LongStream, a streaming geometry foundation model centered on a ``gauge-decoupled" design. It predicts keyframe-relative poses and employs orthogonal scale decoupling. This design systematically eliminates first-frame anchor dependence and effectively mitigates failures in long-sequence extrapolation.
    
    \item We identify attention-sink reliance and KV-cache contamination as the primary causes of long-horizon degradation. Our cache-consistent training scheme and periodic cache refresh stabilize temporal attention and reduce geometric drift.
    
    \item  Experiments across indoor and outdoor benchmarks show that LongStream achieves state-of-the-art performance under a strictly online, future-invisible setting, while maintaining real-time throughput and stable metric scale on long sequences.
\end{itemize}

\section{Related Work}
\label{sec:RW}

\begin{figure*}
  \vspace{-5pt}
  \begin{center}
  \includegraphics[width=\linewidth]{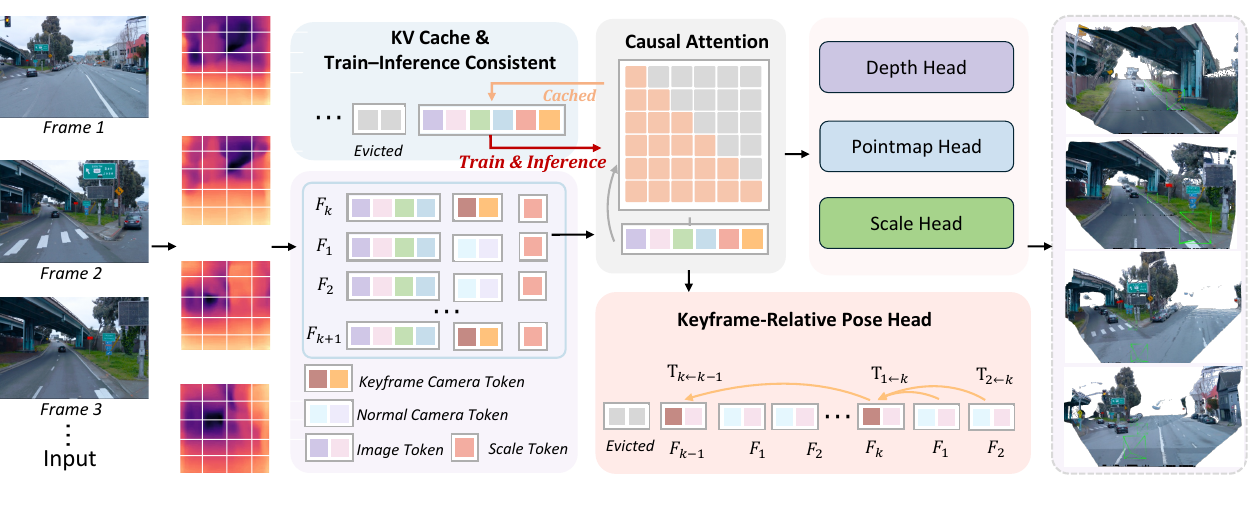}
  \end{center}
  \vspace{-20pt}
  \caption{\textbf{Overview of our proposed LongStream.} Given streaming inputs, patch tokens are extracted by a ViT encoder and augmented with \textit{keyframe}, \textit{normal-frame}, and \textit{scale tokens}. Tokens are fused via causal attention with a shared KV cache, which is consistently used in both training and inference for cache-consistent streaming modeling. The network predicts keyframe-relative poses $\mathbf{T}_{i\leftarrow k}$, depth, pointmap, and global scale, 
  enabling stable metric-scale reconstruction over long sequences. 
  }
   \label{fig:framework}
  \vspace{-15pt}
\end{figure*}

\noindent \textbf{Classical SfM and MVS.}
Structure-from-Motion (SfM) and Multi-View Stereo (MVS) pipelines~\cite{Schonberger_2016_CVPR,schoenberger2016mvs,yao2018mvsnetdepthinferenceunstructured,yao2019recurrent,Galliani_2015_ICCV,fu2022geoneusgeometryconsistentneuralimplicit,cheng2025graphguidedscenereconstructionimages}
reconstruct 3D scenes by modeling geometric relations between image correspondences.
SfM estimates camera poses and sparse structure via feature matching and bundle adjustment~\cite{Schonberger_2016_CVPR,schoenberger2016mvs}, while MVS densifies them with pixel-wise depth from plane sweeping or cost volumes~\cite{yao2018mvsnetdepthinferenceunstructured,yao2019recurrent,Galliani_2015_ICCV,fu2022geoneusgeometryconsistentneuralimplicit,cheng2025graphguidedscenereconstructionimages}.
Despite high accuracy and interpretability, these optimization-heavy pipelines with handcrafted features scale poorly to large or dynamic scenes and are difficult to deploy in real time.

\noindent \textbf{Offline 3D reconstruction.} 
Early end-to-end methods are mainly trained on image pairs~\cite{charatan2024pixelsplat,xu2025freesplatter,zhang2024monst3r,zhang2024gs}. 
Pointmaps offer efficient computation and real-time usage compared to voxel, mesh, or implicit fields~\cite{sitzmann2019deepvoxels,gkioxari2019mesh,park2019deepsdf,mildenhall2021nerf,cheng2025reggsunposedsparseviews}, enabling SLAM and neural rendering~\cite{murai2025mast3r,kerbl20233d,cheng2025unposed3dgsreconstructionprobabilistic,cheng2025outdoormonocularslamglobal,yu2025rgbonlygaussiansplattingslam}. 
DUSt3R~\cite{wang2024dust3r} regresses pointmaps and relative pose from two images without intrinsics, but remains pairwise and requires global alignment. 
MASt3R~\cite{leroy2024grounding} adds dense features and reciprocal matching to improve calibration and matching robustness, yet operates on pairs and needs costly fusion for multi-view scenes. 
VGGT~\cite{wang2025vggt} predicts poses, depth, pointmaps, and tracks from multi-view inputs in a single feed-forward pass, but relies on a fixed reference frame and absolute pose supervision, causing reference and scale biases. 
$\pi^3$~\cite{wang2025pi} removes reference-view bias via permutation-equivariant design and predicts local pointmaps, but outputs remain ambiguous up to a global similarity transform and lack metric scale.

\noindent \textbf{Streaming 3D reconstruction.}
Streaming methods update geometry frame by frame. 
Classical monocular SLAM and learning-based variants~\cite{davison2007monoslam,liu2025slam3r,zhu2024nicer,choy20163d,yu2021pixelnerf,wang2021ibrnet} incrementally recover structure and motion. 
CUT3R~\cite{wang2025continuous} maintains a recurrent state to output metric pointmaps online, but its RNN backbone struggles with long-term dependencies and degrades on long sequences. 
Stream3R~\cite{lan2025stream3r} adopts a causal Transformer with a KV cache, scaling to long streams but suffering attention collapse as cached tokens dominate. 
StreamVGGT~\cite{zhuo2025streaming} adds temporal causal attention and cache updates, with distillation improving consistency, yet long-horizon stability remains difficult due to cache contamination. 
Overall, existing streaming methods degrade noticeably as sequences grow and fail to generalize to much longer streams.

\section{Methodology}
\subsection{Overall}
\label{sec:arch}

We propose LongStream, a \emph{gauge-decoupled} streaming geometry framework that jointly predicts \text{pose}, \text{depth}, and \text{scale} within a unified spatiotemporal Transformer, as shown in Figure~\ref{fig:framework}. 
A ViT encoder produces patch tokens augmented with \emph{Camera}, \emph{Register}, and \emph{Scale} tokens to distinguish keyframe roles. 
These tokens are fused by a causal aggregator with a shared \emph{KV cache}, enabling long-sequence streaming inference. 
For each frame, the model predicts a keyframe-relative pose $\mathbf{T}_{i\leftarrow k}$, depth, a pointmap, and a global scale $s$. 
Training and inference use the same layout, ensuring gauge-decoupled and stable metric-scale reconstruction. We next detail the components of the design.

\subsection{Gauge-Decoupled Formulation}
\label{sec:gauge_decoupling}

We aim to overcome the limitations of existing streaming models that rely on gauge-coupled designs. 
We argue that a robust geometric learning system must remain theoretically invariant to the gauge freedoms, namely arbitrary global coordinates and metric scale. 
To this end, we propose a framework that systematically separates the $SE(3)$ and $Sim(3)$ degrees of freedom.

In the \text{$SE(3)$ gauge}, we discard absolute pose regression and redefine pose learning as 
\text{gauge-invariant}~\cite{ba,8354808} relative pose estimation. 
The learning objective becomes:
\begin{equation}
    \mathbf{T}_{i \leftarrow k} = \mathbf{T}_i \circ \mathbf{T}_k^{-1},
\end{equation}
where $\mathbf{T}_i$ and $\mathbf{T}_k$ denote world-to-camera absolute poses and the $k$-th frame is the preceding keyframe. 
This formulation is mathematically gauge-invariant under any world-frame reparameterization 
$S \in SE(3)$. 
The decoupling design achieves two goals simultaneously: 
it transforms an \text{out-of-distribution} long-range extrapolation problem, caused by large indices~\cite{lan2025stream3r, zhuo2025streaming}, 
into a constant-difficulty \text{in-distribution} local task with bounded index gap $(i-k)$; 
and it mitigates the fixed-anchor bias that contributes to instability in first-frame–anchored causal models.

In the \text{$Sim(3)$ gauge}, we address chaotic scale entanglement with the scale-invariant (SI-Log) philosophy \cite{yang2024depthanythingunleashingpower}.
We use an orthogonal scale learning mechanism, where we separate geometry learning and metric scale estimation at both the architectural and objective levels.
The geometry branch is normalized and supervised with scale-invariant principles~\cite{yang2024depthanythingunleashingpower}. A dedicated scale head then predicts the global scale factor $s$.

\subsection{Network Architecture}
\label{sec:architecture}

Given input images $I_i$, the model predicts, for each frame $i$,
its relative pose with respect to the reference keyframe $k$,
$\mathbf{p}_{i\leftarrow k}=[\mathbf{t}_{i\leftarrow k},\,\mathbf{q}_{i\leftarrow k},\,f_{i\leftarrow k}]$,
where $\mathbf{t}$ is translation, $\mathbf{q}$ a unit quaternion, and $f$ a focal-length offset.
The model also outputs the depth map $D_i$, the corresponding world-coordinate point cloud $X_i$,
a global scale factor $s$, and the frame representation $h_i$:
\begin{equation}
\{h_i, \mathbf{p}_{i\leftarrow k}, D_i, X_i, s\}
= F_\theta(I_i), \qquad i=1,\dots,S.
\end{equation}
where $h_i$ denotes the frame-level feature representation.  

Following the architecture of VGGT-style streaming models~\cite{wang2025vggt,lan2025stream3r,zhuo2025streaming}, the architecture consists of a DINOv2-based~\cite{oquab2024dinov2learningrobustvisual} tokenizer, a causal Transformer aggregator, and task-specific heads for relative pose, depth, pointmap, and scale.  
Geometry is modeled in a keyframe-relative manner: each frame predicts its pose and pointmap with respect to the current keyframe $k$, 
enabling streaming reconstruction without reliance on a fixed first-frame coordinate frame.

For each frame $I_i$, the tokenizer extracts patch features 
$x_i^p \in \mathbb{R}^{P\times C}$ and augments them with camera token and register tokens, 
with distinct tokens for keyframes and non-keyframes.  
A shared Scale Token is added for Sim(3) decoupling.  
All tokens are concatenated into $H^{(0)} \in \mathbb{R}^{B\times S\times(P+T)\times C}$ 
and processed by a stack of Transformer blocks with alternating intra-frame and global attention under a strictly causal mask:
\begin{equation}
H^{(l+1)} = \mathrm{Block}^{(l)}(H^{(l)}, \mathrm{AttnMask}),
\end{equation}
with outputs fed into the pose, depth, and pointmap heads for iterative refinement.

\noindent \textbf{Relative pose head.} It takes both frame and keyframe features from the aggregator and explicitly predicts the relative transformation 
$\mathbf{T}_{i\leftarrow k}$ of the current frame $i$ with respect to its reference keyframe $k$:
\begin{equation}
\mathbf{p}_{i\leftarrow k} = [\mathbf{t}_{i\leftarrow k}, \mathbf{q}_{i\leftarrow k}, f_{i\leftarrow k}],
\end{equation}
where the translation $\mathbf{t}_{i\leftarrow k}\in\mathbb{R}^3$, rotation $\mathbf{q}_{i\leftarrow k}\in\mathbb{H}$ (unit quaternion), 
and focal offset $f_{i\leftarrow k}\in\mathbb{R}^2$ jointly form the relative camera pose:
\begin{equation}
\mathbf{T}_{i\leftarrow k} = \mathbf{T}_i \mathbf{T}_k^{-1}.
\end{equation}
This definition remains invariant under any right-multiplicative transformation of the world coordinate frame, 
thereby removing the dependency on a fixed world anchor.

To ensure that the model learns only the relative relationship between $(i,k)$, we adopt a \text{reference-aware attention} scheme. 
For a non-keyframe $i$ assigned to keyframe $k$, its tokens attend only to tokens from $k$ and from frames between $k$ and $i$ under the causal or window mask. 
Keyframe tokens, in turn, attend only to tokens from the previous keyframe $k\!-\!1$ and from frames between $k\!-\!1$ and $k$, rather than to the entire history. 
After aggregation, we fuse the pose tokens of the current frame and its reference keyframe via concatenation and a linear projection:
\begin{equation}
\mathbf{h}_{\text{fused}} = \text{Proj}([\mathbf{h}_i, \mathbf{h}'_k]).
\end{equation}
Finally, following the design of RAFT~\cite{teed2020raft}, the head employs an \texttt{AdaLN}-modulated Transformer to iteratively predict the relative pose 
$\mathbf{p}_{\text{rel}}=[\mathbf{t},\mathbf{q},f]$ through incremental updates 
$\mathbf{p}^{(t+1)} = \mathbf{p}^{(t)} + \Delta\mathbf{p}^{(t)}$.

\noindent \textbf{Scale head.} To achieve $Sim(3)$ gauge invariance, we design an explicit \text{scale head} that receives a dedicated \textit{Scale Token}, which is similar to the scale token idea in concurrent work MapAnything~\cite{keetha2026mapanythinguniversalfeedforwardmetric}.
It predicts an unconstrained log-scale variable $x_s \in \mathbb{R}$, which is then exponentiated to obtain a strictly positive scale factor:
\begin{equation}
s = \exp(\mathbf{w}^\top \mathbf{h}_{\text{scale}}),
\end{equation}
where $\mathbf{h}_{\text{scale}}$ is the feature of the scale token at the last aggregator layer. 
The scale $s$ affects only translation, depth, and pointmap outputs, 
while rotation and field of view remain unchanged. 
The scale head is trained only on datasets with available ground-truth metric scale.

\noindent \textbf{Depth and pointmap heads.} The \text{depth head} and \text{pointmap head} take the aggregated frame-level and patch-level features to predict, for each frame, 
a depth map $D_i\in\mathbb{R}^{H\times W}$ and the corresponding world-coordinate points $X_i\in\mathbb{R}^{H\times W\times3}$, 
along with per-pixel confidence scores. 
These branches operate jointly with the scale head: geometry is optimized in a \text{scale-invariant} space, 
while the \textit{ScaleToken} independently learns the global scaling factor, 
ensuring full $Sim(3)$ gauge decoupling.

\begin{figure*}
  \vspace{-5pt}
  \begin{center}
  \includegraphics[width=\linewidth]{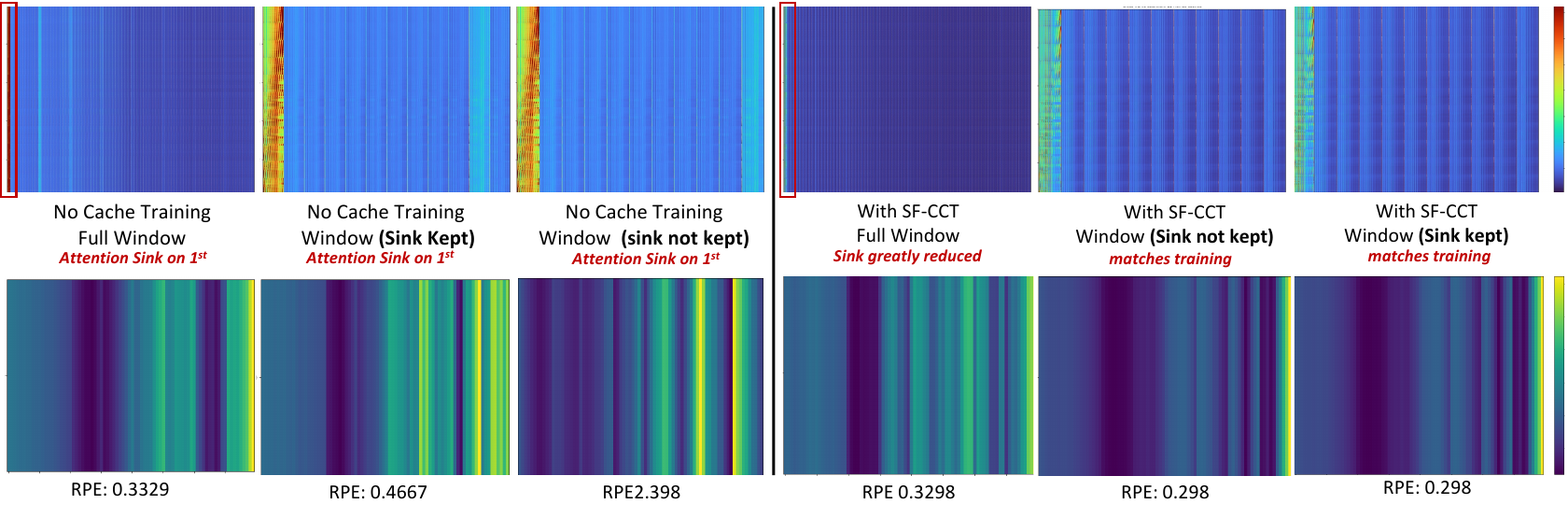}
  \end{center}
 
  \vspace{-20pt}
  \caption{
  \textbf{Cache-consistent training (CCT).}
We show attention maps (top) and Relative Pose Error (RPE) heatmaps (bottom) under different training–inference settings.
Without CCT (left), causal inference develops a strong attention sink; windowed inference either amplifies this sink when it is kept or collapses when it is removed.
With CCT (right), the sink is strongly suppressed in causal mode and likewise suppressed in both windowed modes, yielding stable and best accuracy.
Light blue denotes attention to the keyframe.}
   \label{fig:attention}
  \vspace{-9pt}
\end{figure*}

\subsection{Probabilistic Framework and Loss Functions}
\label{sec:loss}

To implement the gauge-decoupled design within a unified probabilistic framework, 
we formulate the overall objective as maximizing the joint likelihood of geometry, motion, and scale given the input sequence.
Let $I$ denote the image frames, $X$ the 3D pointmap, $D$ the depth, $p$ the relative pose, and $s$ the global scale.
We minimize the \text{Negative Log Posterior}:
\begin{equation}
\mathcal{L}
=\underbrace{\mathcal{L}_{\text{geom}}+\mathcal{L}_{\text{depth}}}_{\text{Geometry \& Depth Likelihood}}
+\underbrace{\mathcal{L}_{\text{pose}}}_{\text{Pose Likelihood}}
+\underbrace{\mathcal{L}_{\text{scale}}}_{\text{Scale Prior}},
\end{equation}
where each term corresponds to a conditional factor in the posterior decomposition
\begin{align}
p(D,X,p,s\mid I)\propto\;&p(D\mid X,I)\cdot p(X\mid p,s,I)\cdot\nonumber\\
&p(p\mid I)\cdot p(s).
\end{align}
This factorization aligns the learning process with the gauge-invariant formulation of $SE(3)$ and $Sim(3)$ introduced in Sec.~\ref{sec:gauge_decoupling}.

\begin{figure*}
\vspace{-5pt}
\begin{center}
\includegraphics[width=0.85\textwidth]{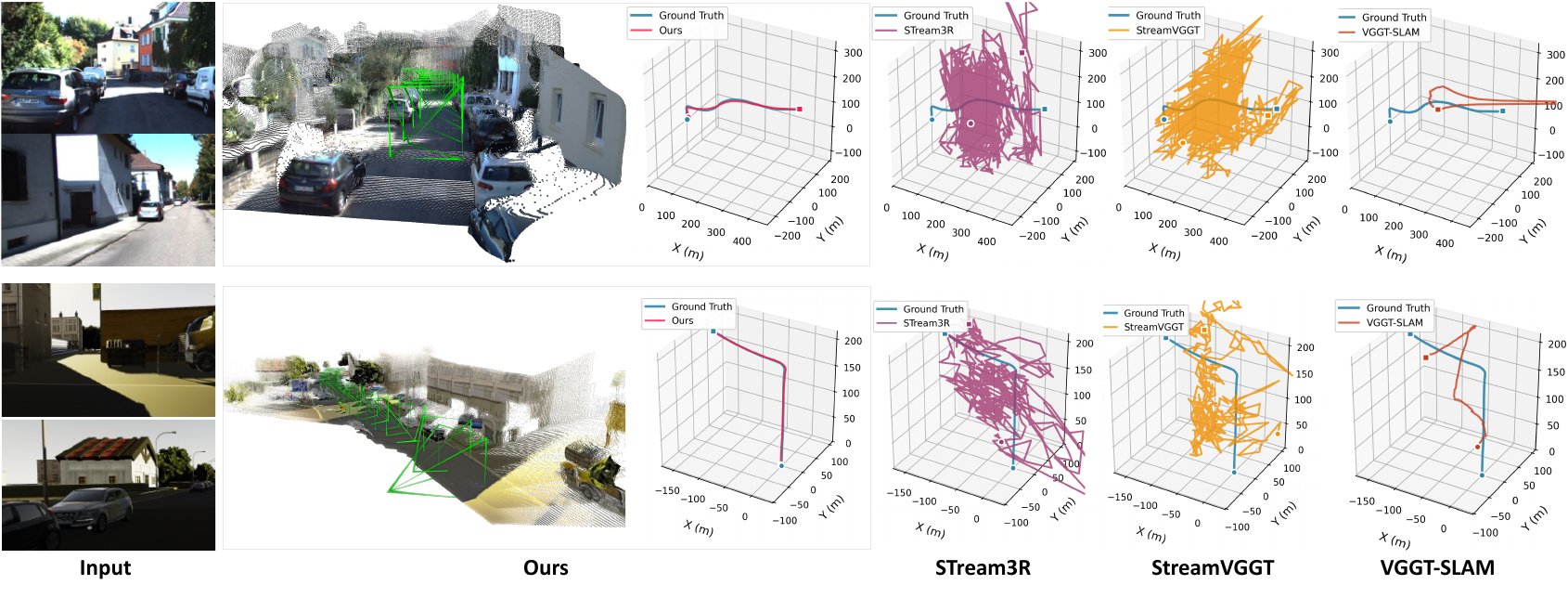}
\end{center}
\vspace{-20pt}
\caption{
\textbf{Qualitative comparison on long-sequence pose estimation.}
We compare LongStream against streaming and SLAM baselines on KITTI and vKITTI sequences spanning several hundred meters. Stream3R and StreamVGGT accumulate drift over long trajectories, and VGGT-SLAM runs out of memory on the second vKITTI sequence. LongStream preserves stable and coherent poses across all scenes, maintaining trajectory continuity even under large loop motions.}
\label{fig:kitti}
\vspace{-5pt}
\end{figure*}

\begin{table*}[t]
\centering
\footnotesize % 使用 \small 或 \footnotesize 代替 \resizebox 来保持字体一致性
\setlength{\tabcolsep}{3pt} % 稍微减少列间距以适应页面宽度 (默认是 6pt)
\begin{tabular}{l ccccc ccccc c |c}
\toprule
\multirow{3}{*}{\textbf{Methods}} & \multicolumn{11}{c}{\textbf{KITTI~\cite{Geiger2012CVPR}} (ATE $\downarrow$)} & \multirow{3}{*}{\textbf{Avg.}} \\
\cmidrule(lr){2-12} % (lr) 表示线条只跨越 2-12 列，并有轻微的左右边距
& \makecell{00 \\ \tiny 4542x, 3.7km} & \makecell{01 \\ \tiny 1101x, 2.5km} & \makecell{02 \\ \tiny 4661x, 5.1km} & \makecell{03 \\ \tiny 801x, 0.6km} & \makecell{04 \\ \tiny 271x, 0.4km} & \makecell{05 \\ \tiny 2761x, 2.2km} & \makecell{06 \\ \tiny 1101x, 1.2km} & \makecell{07 \\ \tiny 1101x, 0.7km} & \makecell{08 \\ \tiny 4071x, 3.2km} & \makecell{09 \\ \tiny 1591x, 1.7km} & \makecell{10 \\ \tiny 1201x, 0.9km} & \\
\midrule
FastVGGT     & *      & 705.39   & *      & 62.38    & 10.27    & 157.74   & 124.43   & 69.27    & *      & 190.10   & 194.75   & 189.29   \\
MASt3R-SLAM  & *      & 530.37   & *      & 18.87    & 88.98    & 159.430   & 92.00    & *      & *   & *       & *   & 177.93   \\
VGGT-SLAM    & *       & 607.16   & *       & 169.83   & 13.12    & *       & *       & *       & *       & *       & *   & 263.37   \\
\midrule
CUT3R        & 185.89   & 651.52   & 296.98   & 148.06   & 22.17    & 155.61   & 132.54   & 77.03    & 238.39   & 205.94   & 193.39   & 209.78   \\
TTT3R        & 190.93   & 546.84   & 218.77   & 105.28   & 11.62    & 153.12   & 132.94   & 70.95    & 180.57   & 211.01   & 133.00   & 177.73   \\
STream3R     & 190.98   & 681.95   & 301.40   & 158.25   & 102.73   & 159.85   & 135.03   & 90.37    & 261.15   & 216.31   & 207.49   & 227.77   \\
StreamVGGT   & 191.93   & 653.06   & 303.35   & 157.50   & 108.24   & 160.46   & 133.71   & 89.00    & 263.95   & 216.69   & 209.80   & 226.15   \\
\midrule % 在 "Ours" 方法前加一条分割线，可以更突出显示
\textbf{Ours} & \textbf{92.55} & \textbf{46.01} & \textbf{134.70} & \textbf{3.81} & \textbf{1.95} & \textbf{84.69} & \textbf{23.12} & \textbf{14.93} & \textbf{62.07} & \textbf{85.61} & \textbf{21.48} & \textbf{51.90} \\
\bottomrule

\end{tabular}

\vspace{-8pt}
\caption{\textbf{Quantitative comparison on the KITTI dataset in terms of ATE.} The upper block lists optimization-based baselines, and the lower block reports streaming methods. Our approach achieves the best accuracy across all sequences.}
\vspace{-10pt}
\label{tab:kitti_result}
\end{table*}

\noindent \textbf{Relative pose loss.}
The relative pose loss $\mathcal{L}_{\text{pose}}$ corresponds to $p(p\mid I)$ and supervises the \texttt{RelPoseHead} output 
$\mathbf{p}_{\text{rel}}=[\mathbf{t},\mathbf{q},f]$ across iterative updates:
\begin{align}
\mathcal{L}_{\text{pose}}
=\sum_{t=1}^{T}\gamma^{t-1}\Big(
&\ell(\hat{q}^{(t)},q_{i\leftarrow k})
+\ell_t(\hat{t}^{(t)},t_{i\leftarrow k}) \nonumber \\
&+\ell(\hat{f}^{(t)},f_{i\leftarrow k})
\Big),
\label{eq:pose_loss}
\end{align}
where $\ell$ denotes a L1 loss and $f_{i\leftarrow k}$ is the focal offset.
To maintain gauge decoupling, the translation term $\ell_t$ is computed in a normalized coordinate space, 
ensuring that translation supervision does not implicitly encode global scale.

\noindent \textbf{Geometry loss.}
Inspired by SI~\cite{yang2024depthanythingunleashingpower}, the geometric loss $\mathcal{L}_{\text{geom}}$ (shape optimization) corresponds to $p(X\mid p,s,I)$ 
and operates in the normalized space:
\begin{equation}
\begin{aligned}
\tilde{X}_{\text{pred}} &= 
\frac{\hat{X}_{\text{raw}}}{\text{Norm}(\hat{X}_{\text{raw}})},\quad
\tilde{X}_{\text{gt}} =
\frac{X}{\text{Norm}(X)},\\[3pt]
\mathcal{L}_{\text{geom}} &= 
\|\tilde{X}_{\text{pred}}-\tilde{X}_{\text{gt}}\|_1.
\end{aligned}
\end{equation}
Normalization removes explicit scale dependency, ensuring $\partial\mathcal{L}/\partial s=0$. 
Hence, $\mathcal{L}_{\text{geom}}$ supervises only the backbone to learn correct 3D structure.

\noindent \textbf{Scale loss.}
The scale loss $\mathcal{L}_{\text{scale}}$ regularizes the global scale $s$.
Since scale is multiplicative, we compare predicted and ground-truth scale in log space to measure relative error and stabilize gradients:
\begin{equation}
\mathcal{L}_{\text{scale}}
=\|\log \hat{s} - \log s_{\text{gt}}\|_1.
\end{equation}
where $s_{\text{gt}}$ is the metric scale computed from calibrated ground-truth depth.
This loss is applied only to metric-calibrated samples; non-metric data are trained using geometry and depth losses alone.

\begin{algorithm}[t]
\caption{Cache-Consistent Training}
\begin{algorithmic}[1]
\State \textbf{Input:} chunks $\{c_1,\dots,c_N\}$, initial cache $\mathrm{KV}^{(0)}=\emptyset$
\For{$i=1$ to $N$}
    \State $(\text{out}_i,\,\mathrm{KV}^{\text{new}}) = \text{model}(c_i,\,\mathrm{KV}^{(i-1)})$
    \State $\mathrm{KV}^{(i)} = \text{trim}(\mathrm{KV}^{\text{new}},\,\text{window\_size})$
\EndFor
\end{algorithmic}
\label{afcct}
\end{algorithm}

\subsection{KV Cache and Train--Inference Consistency}
\label{sec:kv_consistency}

Prior work on streaming Transformers shows that models often rely on an \text{attention sink}~\cite{xiao2024efficientstreaminglanguagemodels,xu2025streamingvlmrealtimeunderstandinginfinite,yang2025longliverealtimeinteractivelong,shin2026motionstreamrealtimevideogeneration} to stabilize attention; without it, sliding past the first frame can trigger \text{model collapse}. 

However, this anchoring creates a fragile reliance on the first frame. Over long sequences, it leads to geometric saturation~\cite{yang2025longliverealtimeinteractivelong}, manifested as unstable attention, keyframe jumps, and growing pose errors.
These issues persist even with relative-pose supervision, indicating that sink anchoring itself imposes an asymmetric positional bias.

We argue that “short-horizon collapse” is not caused directly by removing the sink, but is a symptom of train–inference mismatch.
We therefore introduce Cache-Consistent Training (CCT), introduced in Algorithm~\ref{afcct}, which explicitly passes and trims the KV cache between chunks to align training and inference visibility. 
During training, we remove the constant sink token and use purely causal masking with a sliding window, while \textit{explicitly passing and trimming} the KV cache between chunks so that cache visibility mirrors inference.

As shown in Figure \ref{fig:attention}, CCT makes the attention pattern \text{mathematically equivalent} between chunked training and frame-by-frame inference, forcing the model to operate in a \text{pure} sliding window without a persistent anchor and thereby removing sink dependence.

For ultra-long sequences, accumulated KV still yields long-term memory saturation and geometric drift. 
We thus adopt a \text{periodic cache refresh} that hard-marginalizes stale context by resetting the sink frame and KV cache every $N$ keyframes, akin to state marginalization in SLAM. 

This retains geometric continuity while clearing degraded features, allowing periodic memory resets with no extra compute; because the entire model operates in a keyframe-relative coordinate system, the cache can be refreshed at any keyframe without breaking consistency or degrading accuracy.

Combining CCT with periodic cache refresh yields stable, generalizable streaming over thousands of frames, maintaining consistent geometric accuracy and well-behaved attention distributions.

\section{Experiments}

\begin{figure*}
\vspace{-5pt}
\begin{center}
\includegraphics[width=0.86\textwidth]{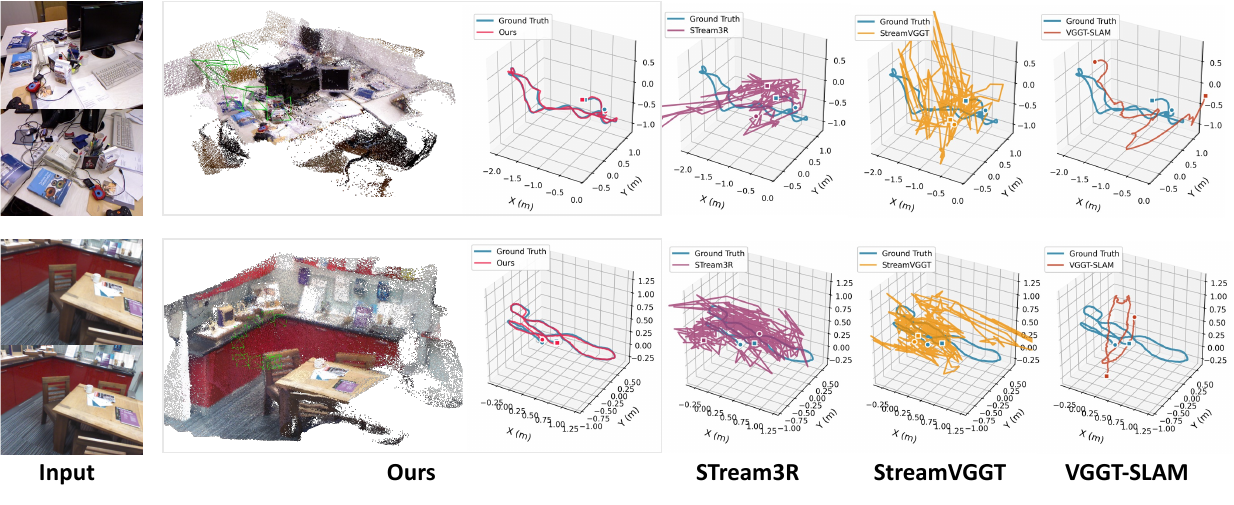}
\end{center}
\vspace{-18pt}

\caption{\textbf{Qualitative comparison on indoor sequences.} We evaluate challenging scenes with strong viewpoint changes, occlusions, and repeated back-tracking. While Stream3R, StreamVGGT, and VGGT-SLAM drift on these highly folded trajectories, LongStream maintains stable poses and consistent 3D structure throughout the sequence. The symbol * denotes runs with out-of-memory errors or more than three tracking failures.}
\label{fig:tum}
\vspace{-10pt}
\end{figure*}

\begin{table}[t]
\centering
\small
\setlength{\tabcolsep}{6pt}
\renewcommand{\arraystretch}{1.2}
\resizebox{0.42\textwidth}{!}{%
\begin{tabular}{lccc}
\toprule
\multirow{2}{*}{\textbf{Methods}} &
\textbf{TUM}~\cite{sturm12iros} &
\textbf{Oxford Spires}~\cite{tao2025oxfordspiresdatasetbenchmarking} &
\textbf{Waymo}~\cite{sun2020scalabilityperceptionautonomousdriving} \\
\cmidrule(lr){2-4}
& ATE $\downarrow$ & ATE $\downarrow$ & ATE $\downarrow$ \\
\midrule
FastVGGT      & 0.418    & 36.577   & 1.281 \\
MASt3R-SLAM   & 0.082  & 37.728   & 7.625 \\
VGGT-SLAM     & 0.123 (0.053$^{\dagger}$)  & 31.003   & 7.431 \\
\midrule
CUT3R         & 0.542  & 32.440   & 9.396 \\
TTT3R         & 0.308  & 36.214   & 3.486 \\
STream3R      & 0.633  & 37.569   & 42.203 \\
StreamVGGT    & 0.627  & 37.255   & 45.101 \\
\midrule
Ours          & \textbf{0.076} & \textbf{19.815} & \textbf{0.737} \\
\bottomrule
\end{tabular}
}
\label{tab:small}
\vspace{-6pt}
\caption{\textbf{Quantitative comparison on TUM~\cite{sturm12iros}, Oxford Spires~\cite{tao2025oxfordspiresdatasetbenchmarking}, and Waymo~\cite{sun2020scalabilityperceptionautonomousdriving}.} Top: optimization-based methods; Bottom: streaming methods. Our method demonstrates robustness on these small-scale trajectories, achieving the best performance across all online benchmarks. $^{\dagger}$ Reported in~\cite{maggio2025vggtslamdensergbslam}.}
\vspace{-8pt}
\label{tab:tum_oxford_waymo}
\end{table}

\begin{table}[t]
\centering
\small
\setlength{\tabcolsep}{1.8pt}
\renewcommand{\arraystretch}{1.2}
\resizebox{0.48\textwidth}{!}{
\begin{tabular}{lccccc|c}
\toprule
\multirow{3}{*}{Methods} &
\multicolumn{6}{c}{vKITTI~\cite{cabon2020virtual} (ATE $\downarrow$)} \\
\cmidrule(lr){2-7}
& \makecell{\footnotesize Scene 01 \\ $447\times, 332m$}
& \makecell{\footnotesize Scene 02 \\ $223\times, 113m$}
& \makecell{\footnotesize Scene 06 \\ $270\times, 51m$}
& \makecell{\footnotesize Scene 18 \\ $339\times, 254m$}
& \makecell{\footnotesize Scene 20 \\ $837\times, 711m$}
& \makecell{\textbf{Avg.}} \\
\midrule
FastVGGT      & 3.435  & 0.311  & 0.120  & 2.050  & 101.667 & 31.427 \\
MASt3R-SLAM   & 83.771 & 20.206 & 3.840  & 68.875 & 231.064 & 98.714 \\
VGGT-SLAM     & 25.128 & 0.237  & 0.281  & 1.641  & 68.840  & 23.667 \\
\midrule
CUT3R         & 50.968 & 29.913 & 0.820  & 29.012 & 127.583 & 55.276 \\
TTT3R         & 29.877 & 11.785 & 0.598  & 7.445  & 71.208  & 28.099 \\
STream3R      & 68.280 & 26.450 & 8.185  & 43.597 & 198.279 & 82.815 \\
StreamVGGT    & 71.616 & 15.349 & 10.274 & 23.900 & 221.407 & 83.916 \\
\midrule
Ours          & \textbf{1.422} & \textbf{0.185} & \textbf{0.303} & \textbf{0.683} & \textbf{4.030}  & \textbf{1.610} \\

\bottomrule
\end{tabular}
}
\label{vkitti}
\vspace{-6pt}
\caption{\textbf{Quantitative comparison on vKITTI.} Top: optimization-based methods; Bottom: streaming methods. Our method achieves the best accuracy across all sequences.}
\vspace{-16pt}
\label{tab:vkitti_results}
\end{table}

\subsection{Implementation Details}
\noindent\textbf{Model configurations.} We initialize LongStream from VGGT and retain its $24$-layer backbone with alternating global and frame-level attention, containing roughly $1.3B$ parameters. 
During training, we use a fixed visibility layout, a consistent sliding window, and a keyframe interval of ten so that each batch contains multiple keyframe transitions.
We optimize with AdamW and cosine decay, using a peak learning rate of $4\times10^{-6}$, and a warmup of $1k$ steps. All images, depths, and pointmaps are resized to a maximum long side of $518$ pixels, with aspect-ratio jittering, interval sampling, and cross-block shuffling.

\noindent\textbf{Training method.}  
We train LongStream with a two-stage schedule: the first stage performs batch-independent training with batch size $22$ for $50k$ iterations over three days on $32$ A100 GPUs. The second stage applies KV-cache–consistent training that matches streaming inference by sampling sequence lengths between $10$ and $80$ with a cache window of $10$. 
Metric scale supervision is used only when calibrated ground truth is available. 
At inference time, LongStream reaches $18$ FPS on a single GPU. Further implementation details are available in the Appendix.

\noindent\textbf{Training data.} LongStream is trained on a multi-domain dataset collection, including 
Kubric~\cite{greff22kubric:}, 
WildRGB~\cite{xia2024rgbd}, 
ScanNet~\cite{dai2017scannet}, 
HyperSim~\cite{hypersim}, 
Mapillary~\cite{MPSD_2020_ECCV}, 
Replica~\cite{straub2019replica}, 
MVS-Synth~\cite{mvssynth}, 
PointOdyssey~\cite{zheng2023point}, 
Virtual KITTI~\cite{cabon2020virtual},
Aria Synthetic Environments~\cite{Pan_2023_ICCV},
Aria Digital Twin~\cite{Pan_2023_ICCV}
Objaverse~\cite{deitke2023objaverse}, Spring~\cite{mehl2023springhighresolutionhighdetaildataset}, and Waymo Open~\cite{sun2020scalabilityperceptionautonomousdriving}. 
BlendedMVS~\cite{yao2020blendedmvs}, Co3Dv2~\cite{reizenstein21common}, MegaDepth~\cite{li2018megadepth}, and DL3DV~\cite{ling2024dl3dv} do not provide metric ground truth and are therefore excluded from scale training, while all other datasets used in training provide metric supervision.

\noindent\textbf{Baselines.}
We compare LongStream with offline transformers, streaming models, and SLAM-style systems. Because both VGGT~\cite{wang2025vggt} and $\pi^3$~\cite{wang2025pi}exceed memory limits on long clips, we report FastVGGT~\cite{shen2025fastvggttrainingfreeaccelerationvisual} as a practical replacement. Streaming baselines include CUT3R~\cite{wang2025continuous}, TTT3R, STream3R~\cite{lan2025stream3r}, and StreamVGGT~\cite{zhuo2025streaming}. MASt3R-SLAM~\cite{murai2025mast3r} and VGGT-SLAM~\cite{maggio2025vggtslamdensergbslam} represent optimization-based SLAM baselines. VGGT-SLAM performs chunk-wise inference over 16/32 frames and thus accesses future frames within each chunk. All methods are evaluated with official default settings under a unified protocol.

%VGGT-SLAM performs windowed multi-frame inference per pass rather than frame-by-frame updates, so we treat it as an offline baseline. All baselines are run with the official default settings under a unified evaluation protocol.

\begin{table}
\centering
\small
\setlength{\tabcolsep}{6pt}
\renewcommand{\arraystretch}{1.2}
\resizebox{0.4\textwidth}{!}{
\begin{tabular}{lcc|cc}
\toprule
\multirow{2}{*}{\textbf{Methods}} &
\multicolumn{2}{c|}{\textbf{7Scenes}} &
\multicolumn{2}{c}{\textbf{TUM}} \\
\cmidrule(lr){2-3} \cmidrule(lr){4-5}
& CD $\downarrow$ & F1@0.25 $\uparrow$
& CD $\downarrow$ & F1@0.25 $\uparrow$ \\
\midrule
FastVGGT      
& 6.373 & \textbf{0.710}
& \underline{0.104} & \underline{0.926} \\
MASt3R-SLAM   
& 5.987 & 0.691
& \textbf{0.057} & \textbf{0.954} \\
VGGT-SLAM     
& 6.306 & \underline{0.696}
& 1.993 & 0.633 \\
\midrule
CUT3R         
& 6.281 & 0.274
& 0.474 & 0.533 \\
TTT3R         
& 6.231 & 0.260
& 0.249 & 0.792 \\
STream3R      
& \underline{6.353} & 0.479
& 1.126 & 0.444 \\
StreamVGGT    
& 6.630 & \underline{0.483}
& 0.680 & 0.402 \\
\midrule
Ours          
& \textbf{2.260} & 0.641
& 0.225 & 0.673 \\
\bottomrule
\end{tabular}
}
\caption{\textbf{Quantitative comparison on 7Scenes and TUM.} CD (lower) and F1@0.25 (higher) are adopted for evaluation. 
Best numbers are in \textbf{bold}; second best are \underline{underlined}.}
\label{tab:7scenes_tum}
\end{table}

\subsection{Quantitative Results}

\noindent\textbf{Camera pose estimation.} 
We evaluate ATE on vKITTI~\cite{cabon2020virtual} (training), Waymo~\cite{sun2020scalabilityperceptionautonomousdriving} (held-out), and the unseen KITTI~\cite{kerbl20233d}, TUM-RGBD~\cite{sturm12iros}, and Oxford Spires~\cite{tao2025oxfordspiresdatasetbenchmarking} datasets. 
As shown in Tables~\ref{tab:kitti_result}--\ref{tab:vkitti_results}, streaming baselines exhibit nonlinear error growth as sequence length increases, likely due to long-horizon history saturation and contamination, whereas LongStream remains stable. Offline models frequently encounter out-of-memory (OOM) and tracking loss on long clips. Despite operating fully streaming, LongStream achieves state-of-the-art accuracy at $18$ FPS and generalizes robustly across environments.

%LongStream outperforms both offline and streaming baselines. While offline models suffer memory overflows (OOM) on long sequences, our method achieves state-of-the-art accuracy at $18$ FPS, demonstrating robust generalization across both large- and small-scale environments.

\noindent\textbf{3D reconstruction.}
We evaluate full-sequence reconstruction on 7Scenes and TUM on test sequences without subsampling, reporting Chamfer Distance and F1@0.25. 
As shown in Table~\ref{tab:7scenes_tum} and Figure~\ref{fig:tum}, LongStream performs competitively with both offline approaches on both benchmarks. 
The small spatial extent and dense frame coverage of these datasets lead to largely saturated metrics. 
For 7Scenes, failure cases of some methods under all-frame evaluation produce CD values several orders of magnitude larger on certain scenes, leading to an inflated mean CD.

\noindent\textbf{Scale estimation.} We evaluate the accuracy of the recovered metric scale on vKITTI. 
LongStream produces a stable scale estimate across the entire sequence, achieving a scale ratio of $0.9905$ with respect to ground truth. 

\begin{table}
\centering
\small
\setlength{\tabcolsep}{6pt}
\resizebox{0.47\textwidth}{!}{
\renewcommand{\arraystretch}{1.2}
\begin{tabular}{lccc|cccc}
\toprule
\textbf{RelPose} & \textbf{Scale Head} & \textbf{CCT} & \textbf{Cache Refresh} &
\textbf{ATE $\downarrow$} & \textbf{RPE $\downarrow$} & \textbf{Scale Err. $\downarrow$} \\
\midrule
\xmark & \xmark & \xmark & \xmark & 8.043 & 2.207 & - \\
\cmark & \xmark & \xmark & \xmark & 2.819 & 0.750 & -\\
\cmark & \cmark & \xmark & \xmark & 2.645 & 0.484 & 0.010 \\
\cmark & \cmark & \cmark & \xmark & 0.984 & 0.454 & 0.032 \\
\cmark & \cmark & \cmark & \cmark & 0.115 & 0.126 & 0.035 \\
\bottomrule
\end{tabular}
}
\caption{\textbf{Ablation study on RelPose, Scale head, CCT, and cache refresh}. Green indicates enabled, red indicates disabled. Rows 2 and 3 ATE gap is caused by a few large trajectory outliers. Scale Error reports absolute scale deviation; lower is better.}
\vspace{-10pt}
\label{tab:ablation_relpose_kvcache}
\end{table}

\subsection{Qualitative Results}

Figures~\ref{fig:kitti} and \ref{fig:tum} visualize trajectories on kilometer-level and room-scale sequences, confirming stable pose prediction under both large and small spatial extents. 
In outdoor settings (Figure~\ref{fig:kitti}), existing streaming methods such as STream3R and StreamVGGT suffer from accumulated drift over long trajectories, while the optimization-based VGGT-SLAM encounters memory limitations and tracking loss on longer sequences. 
In contrast, LongStream preserves trajectory continuity and metric accuracy across several hundred meters, successfully closing large loops without explicit loop closure modules.

Similarly, in indoor environments (Figure~\ref{fig:tum}), the model demonstrates robustness against highly folded camera trajectories characterized by strong viewpoint changes, occlusions, and repeated back-tracking. 
Where baselines tend to produce unstable or drift-prone poses under these erratic motion patterns, LongStream maintains a coherent global trajectory. 

\subsection{Ablation Study}

We conduct ablations on a single vKITTI sequence to validate LongStream’s four core components:
the keyframe-relative pose head, the scale branch, cache-consistent training (CCT), and periodic cache
refresh. As shown in Table~\ref{tab:ablation_relpose_kvcache}, combining all modules reduces ATE from
$8.043$ to $0.115$, nearly two orders of magnitude.

\noindent\textbf{Gauge-decoupled pose and scale.}
Switching from absolute pose regression to our gauge-decoupled formulation provides the largest gain
(Row 1 → Row 2), confirming that separating local geometry from global coordinates is essential for
generalizing beyond the training window. The scale branch is required for metric consistency; removing
it prevents the model from producing a stable global scale across the sequence.

\noindent\textbf{Temporal cache consistency.}
 The periodic cache refresh prevents long-term memory saturation in infinite streams, while the dedicated scale branch is essential for maintaining metric consistency.
 
Combining all components reduces the final ATE by nearly two orders of magnitude relative to the baseline. More detailed analyses are provided
in the Appendix.

\section{Conclusion}
LongStream delivers stable, metric-scale reconstruction over ultra-long sequences, overcoming the drift and extrapolation failures of existing methods. Its gauge-decoupled pose design and cache-consistent training preserve consistent geometry and scale across thousands of frames, achieving strong accuracy and real-time performance on diverse indoor and outdoor datasets. 

\noindent\textbf{Limitations.} The model still assumes a largely static world, relies on a heuristic keyframe schedule, and shows mild degradation in pointmap consistency over very long windows. These limitations suggest clear directions for improving robustness and generality in future work. 
\clearpage
\section{Acknowledgments}
This work is supported by the National Natural Science Foundation of China (No. 62406267), Guangdong Provincial Project (No. 2024QN11X072), Guangzhou-HKUST(GZ) Joint Funding Program (No. 2025A03J3956) and Guangzhou Municipal Education Project (No. 2024312122).

{
    \small
    \bibliographystyle{unsrt}
    \bibliography{main}
}
\clearpage
\clearpage
\setcounter{page}{1}
\maketitlesupplementary

\section{Gauge Invariance of Relative Pose and Scale}

This appendix provides concise proofs that (1) the keyframe–relative pose used in \emph{LongStream} is strictly invariant to the choice of global coordinate frame, and (2) the geometry and scale objectives are orthogonally decoupled.

% ----------------------------
\subsection{SE(3) Gauge Invariance of Keyframe Relative Pose}
% ----------------------------

We show that our learning target
\begin{equation}
\mathbf{T}_{i \leftarrow k} = \mathbf{T}_i\,\mathbf{T}_k^{-1},
\end{equation}
is invariant under any global $SE(3)$ gauge transformation. This guarantees that training is not affected by arbitrary choices of world coordinates.

\paragraph{Gauge transformation.}
Let $\mathbf{G} \in SE(3)$ re-parameterize the world frame $\mathcal{W}$ into $\mathcal{W}'$. For any 3D point $\mathbf{x}$:
\begin{equation}
\mathbf{x}_{\mathcal{W}'} = \mathbf{G}\,\mathbf{x}_{\mathcal{W}}.
\end{equation}

\paragraph{Transformation of absolute pose.}
For a world-to-camera pose $\mathbf{T}$, the corresponding pose in $\mathcal{W}'$ is
\begin{equation}
\mathbf{T}' = \mathbf{T}\,\mathbf{G}^{-1}.
\label{eq:T-transform}
\end{equation}
This follows from enforcing that camera-frame coordinates remain unchanged.

\paragraph{Invariance of keyframe–relative pose.}
We apply Equation \eqref{eq:T-transform} to frames $i$ and $k$:
\begin{equation}
\mathbf{T}'_i = \mathbf{T}_i\mathbf{G}^{-1},
\qquad
\mathbf{T}'_k = \mathbf{T}_k\mathbf{G}^{-1}.
\end{equation}
Then the relative pose in $\mathcal{W}'$ becomes
\begin{equation}
\begin{aligned}
\mathbf{T}'_{i \leftarrow k}
&= \mathbf{T}'_i(\mathbf{T}'_k)^{-1} \\
&= (\mathbf{T}_i\mathbf{G}^{-1})(\mathbf{T}_k\mathbf{G}^{-1})^{-1} \\
&= \mathbf{T}_i\,(\mathbf{G}^{-1}\mathbf{G})\,\mathbf{T}_k^{-1}
= \mathbf{T}_i\mathbf{T}_k^{-1}.
\end{aligned}
\end{equation}
Thus,
\begin{equation}
\mathbf{T}'_{i \leftarrow k} = \mathbf{T}_{i \leftarrow k},
\end{equation}
showing that the target is strictly $SE(3)$ gauge-invariant.

% ----------------------------
\subsection{Sim(3) Orthogonal Decoupling of Scale}
% ----------------------------

We now show that our normalized geometry objective is independent of the global scale factor, ensuring that shape and scale are optimized through separate gradient paths.

Let the predicted metric point cloud be
\begin{equation}
\hat{\mathbf{X}} = s\,\hat{\mathbf{X}}_{\mathrm{raw}},
\end{equation}
where $s>0$ is the global scale predicted by the scale head.

Let $\mathrm{Norm}(\cdot)$ be homogeneous of degree one:
\begin{equation}
\mathrm{Norm}(\alpha\mathbf{X}) = \alpha\,\mathrm{Norm}(\mathbf{X}).
\end{equation}

The normalized prediction used in the geometry loss is
\begin{equation}
\tilde{\mathbf{X}}_{\mathrm{pred}}
= \frac{\hat{\mathbf{X}}}{\mathrm{Norm}(\hat{\mathbf{X}})}
= \frac{s\hat{\mathbf{X}}_{\mathrm{raw}}}
       {s\,\mathrm{Norm}(\hat{\mathbf{X}}_{\mathrm{raw}})}
= \frac{\hat{\mathbf{X}}_{\mathrm{raw}}}
       {\mathrm{Norm}(\hat{\mathbf{X}}_{\mathrm{raw}})}.
\end{equation}

Hence the geometry loss
\begin{equation}
\ell_{\mathrm{geom}}
= \big\|\tilde{\mathbf{X}}_{\mathrm{pred}}
      - \tilde{\mathbf{X}}_{\mathrm{gt}}\big\|_1
\end{equation}
is independent of $s$, and thus
\begin{equation}
\frac{\partial \ell_{\mathrm{geom}}}{\partial s} = 0.
\end{equation}

This confirms that global scale is fully decoupled from shape optimization, and is learned solely through the dedicated scale objective.

\medskip

In summary, keyframe–relative poses provide strict $SE(3)$ gauge invariance, while normalized geometry ensures $Sim(3)$ scale orthogonality. Together they yield a principled gauge-consistent training objective for long-sequence streaming reconstruction.

\begin{figure*}
  \vspace{-5pt}
  \begin{center}
  \includegraphics[width=\linewidth]{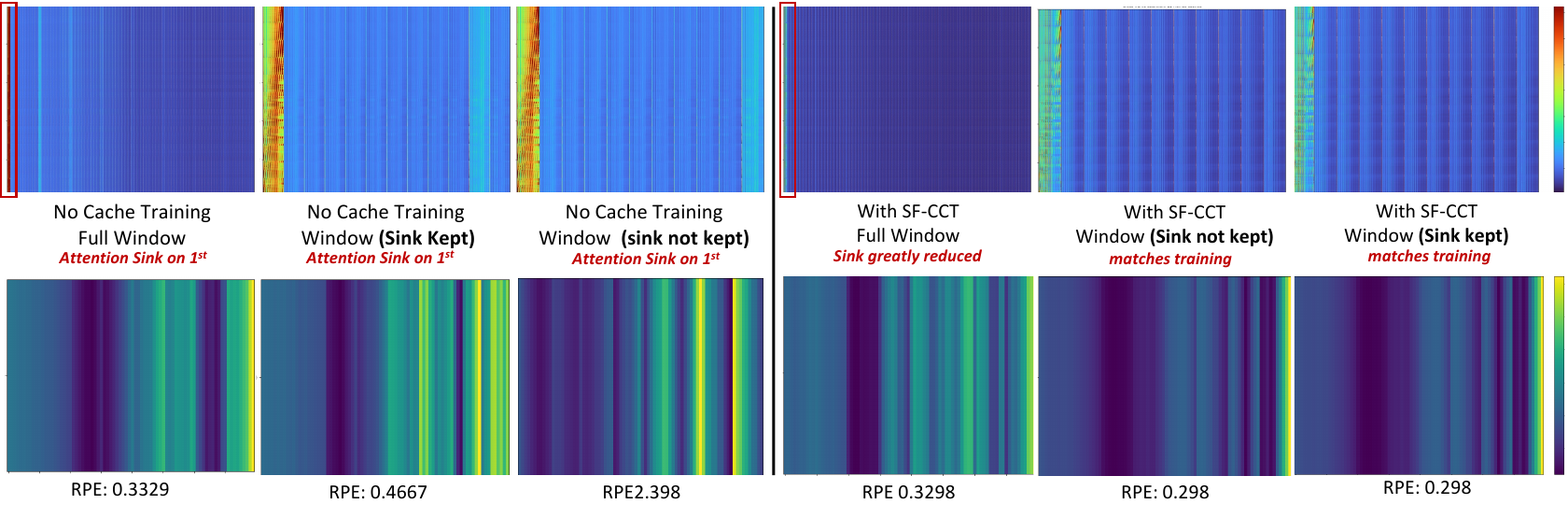}
  \end{center}
 
  \vspace{-20pt}
  \caption{
  \textbf{Cache-consistent training (CCT).}
We show attention maps (top) and Relative Pose Error (RPE) heatmaps (bottom) under different training–inference settings.
Without CCT (left), causal inference develops a strong attention sink; windowed inference either amplifies this sink when it is kept or collapses when it is removed.
With CCT (right), the sink is strongly suppressed in causal mode and likewise suppressed in both windowed modes, yielding stable and best accuracy.
Light blue denotes attention to the keyframe.}
   \label{sup:attention}
  \vspace{-9pt}
\end{figure*}
\section{Additional Attention Visualization Analysis}
\label{si[]:appendix_attention}

As shown in Figure~\ref{sup:attention}, we visualize \emph{frame-level} attention to analyze how the model distributes focus over historical frames during streaming inference. Token–token attention is aggregated into an $S \times S$ frame–frame matrix by summing over target-frame tokens and averaging over source-frame tokens. The causal full-window view contains up to $80$ visible frames, while the sliding-window view contains only $10$, which is reflected in the visualization.

The batch-trained baseline exhibits a clear temporal bias: the model assigns disproportionately high attention to the first frame (the ``sink'') and to more distant frames, while under-attending the recent frames that are most relevant for local geometric consistency. Intuitively, a geometry model should primarily rely on temporally adjacent frames; however, this imbalance causes rapid growth in RPE and unstable long-range predictions. In windowed inference, retaining the sink yields accelerated degradation, whereas removing it leads to collapse, indicating that the baseline is strongly dependent on the initial frame.

With our cache-consistent KV-cache training (CCT), the attention distribution becomes more balanced. The model reduces its reliance on the first frame and allocates relatively more attention to nearby frames, resulting in more stable behavior across both full-window and sliding-window inference. Nonetheless, as sequence length approaches $\sim 80$ frames, we still observe a gradual shift of attention toward earlier history, consistent with cache saturation effects.

Overall, these visualizations highlight the underlying mechanism of long-sequence degradation: baseline models develop a strong first-frame attraction and long-range bias, while CCT encourages attention patterns that better align with temporal geometric coherence.
\section{Long-Sequence Stability Analysis}
Table~\ref{tab:ate_520018670} reports long-sequence results on Waymo \#520018670 (135 m) and KITTI \#03 (561 m). Streaming methods, including baselines and LongStream without refresh, exhibit non-linear error growth as sequence length increases. This suggests that under strictly-online constraints, longer histories or revisiting do not necessarily improve performance and may instead amplify long-horizon effects as the sequence grows. In contrast, LongStream remains stable over long sequences by removing first-frame anchoring and mitigating long-history effects through cache-consistent training with periodic cache refresh.

\begin{table}
\centering
\small
\setlength{\tabcolsep}{3.5pt}
\renewcommand{\arraystretch}{0.92}
\begin{tabular}{lcccc|cc}
\toprule
Method & 15x & 30x & 60x & 199x & 30x & 801x\\
\midrule
CUT3R      & 0.159 & 1.421 & 2.505  & 8.591 & 3.867 & 148.06\\
TTT3R      & 0.153 & 0.873 & 1.127  & 5.505 & 1.957 & 105.28\\
Stream3R   & 0.181 & 1.329 & 2.998 & 21.440 & 3.412 & 158.25\\
% StreamVGGT & 0.192 & 1.418 & 3.295 & 33.292 & 157.50\\
VGGT-SLAM  & 0.172 & 0.518 & 0.992 & 3.740 & 0.929 & 169.83\\
\midrule
Ours (w/o SW) & \textbf{0.151} & 0.192 & 0.477  & 1.699 & 0.347 & 20.83\\
Ours       & \textbf{0.151} & \textbf{0.183} & \textbf{0.343}  & \textbf{0.723} & \textbf{0.164} & \textbf{3.81} \\
\bottomrule
\end{tabular}
\caption{ATE (m) as sequence length increases on Waymo \#520018670 (Left, 135 m) and KITTI \#03 (Right, 561 m). w/o SW denotes the variant without cache refresh and sliding window. }
\label{tab:ate_520018670}
\end{table}

\section{Additional Hyperparameter Analysis}
\label{sec:supp_ablation}

In this section, we provide detailed ablation studies on hyperparameters. These experiments were conducted on the vKITTI dataset to validate our design choices.

\subsection{Impact of Keyframe Interval}
We first examine the sensitivity of the model to the keyframe interval $N$. As presented in Table \ref{tab:supp_interval}, setting an extremely short interval such as $N=1$ degenerates the system into frame-to-frame tracking, leading to rapid error accumulation. Conversely, extending the interval to $15$ also degrades performance. 
This happens because the training chunk is fixed at 22 frames. With such sparse keyframe switches, the model receives too few supervision signals. It cannot reliably learn the switching behaviour.
\begin{table}
    \centering
    \small
    \setlength{\tabcolsep}{12pt}
    \begin{tabular}{ccc}
        \toprule
        Interval $N$ & ATE $\downarrow$ & RPE $\downarrow$ \\
        \midrule
        1  & 4.047 & 0.565 \\
        3  & 3.384 & 0.514 \\
        8  & 0.122 & 0.131 \\
        10 & \textbf{0.115} & \textbf{0.126} \\
        15 & 1.398 & 0.412 \\
        \bottomrule
    \end{tabular}
    \caption{\textbf{Effect of Keyframe Interval.} $N=10$ yields the best trade-off between drift accumulation and training dynamics.}
        \label{tab:supp_interval}
\end{table}

\subsection{Impact of Cache Window Size}
We further investigate the influence of the cache window size $W$. As shown in Table \ref{tab:supp_window}, while a window size of $10$ is sufficient to maintain context, increasing it to $30$ significantly impairs accuracy with the ATE rising to $0.516$. This empirical evidence supports our theory of ``geometric saturation'' where an excessively long history cache accumulates outdated features that pollute the attention mechanism. Thus, a window size of $10$ is adopted to minimize computational cost while preventing long-term drift.

\begin{table}
    \centering
    \small
    \setlength{\tabcolsep}{12pt}
    \begin{tabular}{ccc}
        \toprule
        Window $W$ & ATE $\downarrow$ & RPE $\downarrow$ \\
        \midrule
        10 & \textbf{0.115} & \textbf{0.126} \\
        20 & 0.119 & 0.129 \\
        30 & 0.516 & 0.293 \\
        \bottomrule
    \end{tabular}
    \caption{\textbf{Effect of Cache Window Size.} $W=10$ prevents geometric saturation while maintaining sufficient context.}
        \label{tab:supp_window}
\end{table}

\begin{figure}
  \vspace{-5pt}
  \begin{center}
  \includegraphics[width=\linewidth]{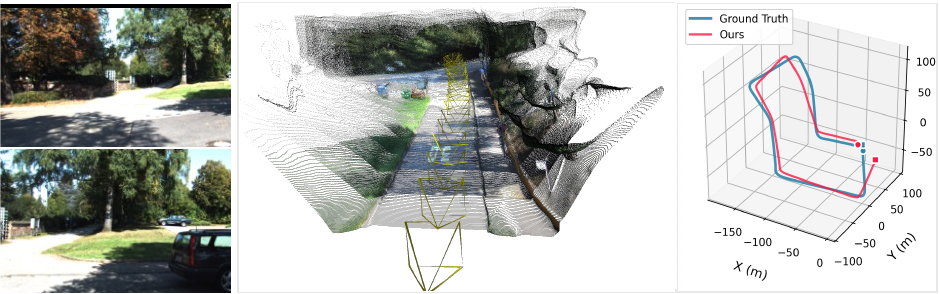}
  \end{center}
 
  \vspace{-8pt}
  \caption{Without loop-closure correction, LongStream shows mild drift when revisiting the same place. Adding online loop-closure cues is a promising direction for improving global consistency.}
   \label{sup:add}
  \vspace{-9pt}
\end{figure}

\section{Additional Limitation}
As illustrated in Figure~\ref{sup:add}, LongStream does not perform explicit loop-closure optimization, and therefore does not benefit from the strong trajectory correction achievable in offline global bundle adjustment. While the proposed relative pose formulation and cache-consistent training already provide stable drift behavior over long horizons, incorporating lightweight online loop-closure cues may further improve global consistency, especially in large loops. We leave this as future work.

% WARNING: do not forget to delete the supplementary pages from your submission 
% \input{sec/X_suppl}

\end{document}